\documentclass[letterpaper, 10 pt, conference]{ieeeconf}  
\usepackage{graphicx}
\usepackage{url}

\IEEEoverridecommandlockouts                              

\title{\LARGE \bf
Next-Gen Museum Guides: 
Autonomous Navigation \\ and Visitor Interaction with an Agentic Robot
}

\author{Luca Garello$^{1}$, Francesca Cocchella$^{1}$, Alessandra Sciutti$^{1}$,  Manuel Catalano$^{2}$, Francesco Rea$^{1}$
\thanks{$^{1}$Member of the Cognitive Architecture for Collaborative Technologies (CONTACT) unit, Italian Institute of Technology}
\thanks{$^{2}$Member of the Soft Robotics for Human Cooperation and Rehabilitation
(SOFTBOTS) unit, Italian Institute of Technology}}

\begin{document}

\maketitle
\thispagestyle{empty}
\pagestyle{empty}

\begin{abstract}
Autonomous robots are increasingly being tested into public spaces to enhance user experiences, particularly in cultural and educational settings. This paper presents the design, implementation, and evaluation of the autonomous museum guide robot Alter-Ego equipped with advanced navigation and interactive capabilities. The robot leverages state-of-the-art Large Language Models (LLMs) to provide real-time, context-aware question-and-answer (Q\&A) interactions, allowing visitors to engage in conversations about exhibits. It also employs robust simultaneous localization and mapping (SLAM) techniques, enabling seamless navigation through museum spaces and route adaptation based on user requests.
The system was tested in a real museum environment with 34 participants, combining qualitative analysis of visitor-robot conversations and quantitative analysis of pre and post interaction surveys. Results showed that the robot was generally well-received and contributed to an engaging museum experience, despite some limitations in comprehension and responsiveness. This study sheds light on HRI in cultural spaces, highlighting not only the potential of AI-driven robotics to support accessibility and knowledge acquisition, but also the current limitations and challenges of deploying such technologies in complex, real-world environments.

\end{abstract}

\section{INTRODUCTION}

Museums serve as vital centers for cultural preservation and education, offering visitors a rich and immersive learning experience. However, navigating large museum spaces and accessing detailed information about exhibits can be challenging, especially for first-time visitors. Traditional methods, such as static signage and audio guides, often provide limited interactivity and fail to adapt to individual visitor preferences. To address these limitations, this paper presents an autonomous museum guide robot that enhances visitor engagement through intelligent navigation and interactive question-and-answer (Q\&A) sessions.
Recent advancements in artificial intelligence (AI) and robotics have enabled the development of autonomous systems capable of understanding and responding to human queries in natural language. Large Language Models (LLMs) have revolutionized conversational AI, allowing robots to provide informative, context-aware responses while adapting to diverse visitor inquiries [1], [2]. Additionally, simultaneous localization and mapping (SLAM) techniques have significantly improved robotic navigation in dynamic indoor environments, making autonomous movement through complex spaces more reliable and efficient [3], [4]. By combining these technologies, the proposed museum guide robot can dynamically interact with visitors and provide personalized tours based on user requests.
This research focuses on the design, implementation, and evaluation of a fully autonomous museum guide robot that integrates LLM-powered dialogue systems with advanced navigation capabilities. The robot can autonomously lead visitors through designated routes, answer questions about exhibits in real time, and personalize tours based on user interests. Through real-world deployment and testing in a museum setting, we assess the system’s performance in terms of navigation efficiency, response accuracy, and user satisfaction.
The remainder of this paper is structured as follows: Section 2 reviews related work on robotic museum guides and conversational AI. Section 3 details the system architecture, including the navigation framework and language processing components. Section 4 presents experimental results from real-world trials. Section 5 discusses key findings, challenges, and future improvements. Finally, Section 6 concludes the paper, highlighting the broader implications of autonomous guide robots in cultural institutions.

\section{Related Works}

Robotic museum guides have evolved significantly, with early systems like RoboX, CiceRobot and Mobot pioneering the use of mobile robots for cultural engagement through multimodal interaction and adaptive tour strategies \cite{chella2007cicerobot}\cite{siegwart2003robox}\cite{1249720}. More recent deployments, including humanoid robots such as Pepper or R1, have sought to increase visitor engagement via social behaviors and interactive tours \cite{kanda2004interactive}\cite{rosa2024tour}. Despite these advancements, a common thread persists across many of these systems: they often operate based on predefined scripts, depend on manual oversight for navigation tasks, or lack the capacity for meaningful, context-aware dialogue.

While prior research has achieved notable progress in autonomous navigation, the conversational layer has frequently been constrained to limited question-answer pairs or preprogrammed speech modules \cite{gasteiger2021deploying}.

Our approach directly addresses these limitations by integrating large language model–powered dialogue with fully autonomous navigation. Unlike earlier systems that treat conversation and movement as separate modules or require operator support, our system enables fluid, natural interactions while navigating independently. 
This tight coupling between language and mobility allows for more intuitive, goal-driven museum exploration.

\newpage

\begin{figure}[htbp] 
    \centering  
    \includegraphics[width=0.5\textwidth, scale=0.10]{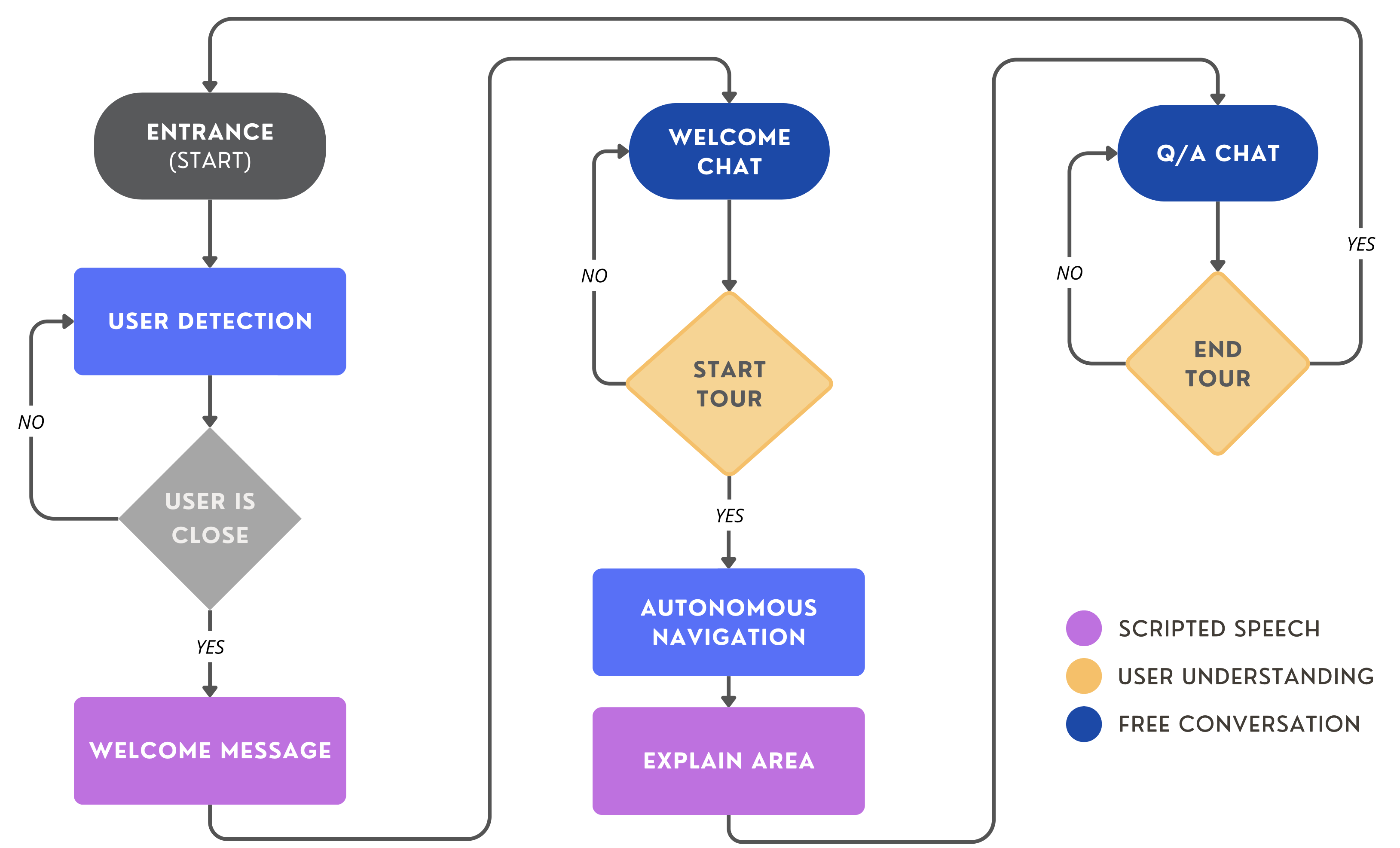}  
    \caption{Simplified flow chart showing the full cycle of a robot guided tour.}  
    \label{fig:robot-tour}  
\end{figure}

\section{System Architecture}
\subsection{Robot}
ALTER-EGO is a dual-arm mobile robot with a functionally anthropomorphic upper body and variable-stiffness actuation for safe human–robot interaction \cite{lentini2019alter}. It features autonomous navigation using Lidar-based SLAM, suitable for dynamic environments. For natural interaction, it integrates a language interface enhanced with a Jabra Speak 510 unit, improving speech clarity and output. 

\subsection{Navigation}
The navigation and localization architecture integrates ROS Hector SLAM for initial map generation, AMCL for adaptive localization, and Move Base for trajectory planning/obstacle avoidance, enhanced by a custom module that enables contextual artwork notifications based on spatial proximity and orientation. This allows the robot to be situationally aware during navigation and trigger pre-recorded utterances such as “We are now passing by the 'Sails' area, where you can see...”.

\subsection{Visitor Recognition and Interaction Initialization}
At the beginning of each interaction, the robot detects the faces of visitors using YOLOv10-n, a state-of-the-art object detection model. The system continuously scans its surroundings for human faces. When a detected face reaches a size threshold (indicating proximity), the robot assumes that the visitor is close enough for interaction. At this point, it greets the visitor, introduces itself, and asks whether they would like to start a guided tour. 
If the visitor agrees, the robot initiates its navigation system using the go\_to() function (see next section). If the visitor declines or does not respond, the robot waits for interaction. If there is no verbal response for more than 120 seconds, the robot assumes disengagement and waits for the next user.

\subsection{Conversation and Function Calling}
User speech was transcribed using Google’s Speech-to-Text API. The robot’s conversational behavior was powered by OpenAI’s GPT-4o mini model, which generated responses dynamically based on visitor queries, the museum context, and the robot’s current location. The LLM is augmented with function-calling capabilities, allowing it to execute actions whenever necessary.
The callable actions are:

\begin{itemize}
    \item go\_to(destination): Moves the robot to the specified exhibit area. The destination is inferred by the LLM among the list of available areas specified in the prompt. This function is only called after the user and the robot agree on the next area to visit. (e.g. "Do you want me to take you to the "Sails" area?")

    \item end\_tour(): if a visitor expresses a desire to stop the tour or there is no verbal response for more than 120 seconds, the LLM can trigger end\_tour(), ensuring a natural and seamless user experience.  This function ends the tour and returns the robot to the entrance.

\end{itemize}

\subsection{Dynamic Prompt Engineering}
To ensure high-quality responses and contextual awareness, the LLM operates on a dynamic prompt structured as follows:

\begin{enumerate}
    \item \textbf{General Robot Information:}
        Identity and purpose of the robot.
        Interaction capabilities (navigation, Q\&A).
        \item \textbf{Current Location:}
        Info about the current location of the robot (e.g. Entrance, Ports of Europe, etc.)
    \item \textbf{Progress of the visit:}
        List of areas already explored + List of areas not yet explored.
    \item \textbf{Knowledge base:}
        Contextual info about the artworks, this includes info about the authors and the their relative position w.r.t. the robot.
        \item \textbf{Chat history:} List of messages exchanged with the user.
\end{enumerate}

At each transition between areas the prompt is dynamically updated in order to adapt the knowledge base of the robot. This dynamic prompt template allows to interrogate the llm with a limited number of input tokens, limiting the operational cost of the robot and reducing the Input/Output delay with the LLM's server.

\subsection{Structured Tour and Autonomous Navigation}
At the entrance, the robot greets visitors and offers to start the tour. It provides brief introductions at each exhibit and initially guides the visitor through the first two areas. After that, the visitor can freely explore, ask questions, or end the tour, while the robot tracks visited areas and offers guidance as needed.
An overview of the interaction flow is presented in Figure \ref{fig:robot-tour}.
A summary of the system behavior is shown in Table \ref{tab:system_behavior}.




\begin{table}[h]
\centering
\begin{tabular}{|l|p{3.8cm}|}
\hline
\textbf{Condition} & \textbf{Robot Behavior} \\
\hline
Visitor detected & Greets and introduces itself. \\
\hline
Visitor agrees to tour & Calls \texttt{go\_to(1st area)}. \\
\hline
Visitor @ 1st area & Calls \texttt{go\_to(2nd area)}. \\
\hline
Visitor asks to visit a specific area & Calls \texttt{go\_to(desired area)} \\
\hline
No interaction for 120 seconds & Calls \texttt{end\_tour()} \\
\hline
Visitor requests to end tour & Calls \texttt{end\_tour()} \\
\hline
\end{tabular}
\caption{Summary of System Behavior}
\label{tab:system_behavior}
\end{table}

\newpage

\section{Experiment}

\begin{figure}[htbp] 
    \centering  
    \includegraphics[width=0.49\textwidth, scale=0.10]{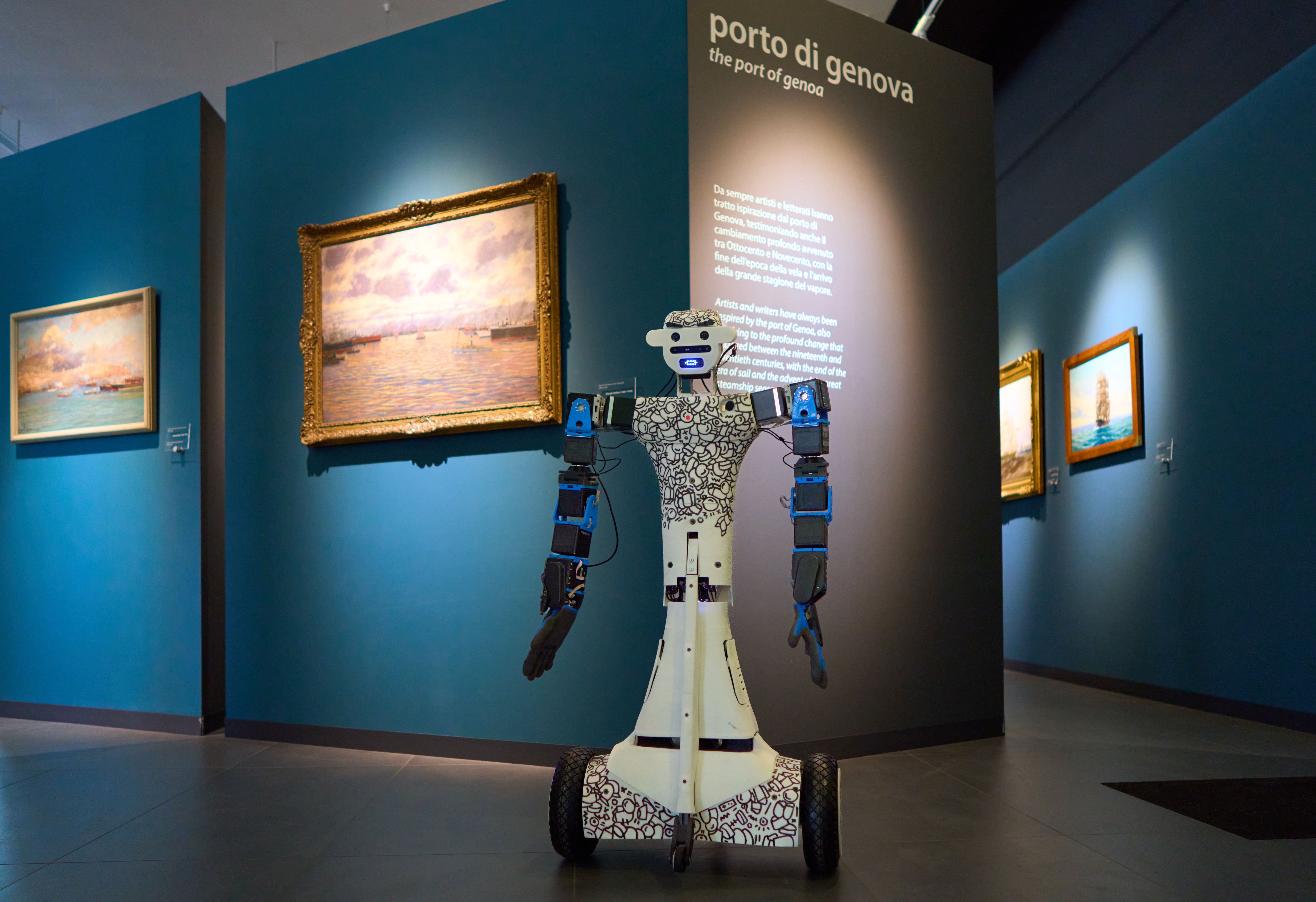}  
    \caption{Alter-Ego robot in front of one area of the tour.}  
    \label{fig:robot}  
\end{figure}

\subsection{Scenario}
The experiment was conducted in a maritime museum featuring exhibitions and interactive installations on the history of navigation. The robot was positioned on a floor showing a collection of maritime paintings. 

\subsection{Research Question and Hypotheses}
This study investigates how visitors perceive a humanoid robot functioning as a museum guide, and whether such perceptions are influenced by interaction and contextual framing.
Based on this, our Research Question (\textbf{RQ}) is: how do people perceive an autonomous museum guide robot in a real-life scenario?
\\Moreover, since we assumed that the interaction with the robot and the expectations about it may influence the robot's perception,we formulated the following hypotheses:
\begin{itemize}
   \item \textbf{H1}: After the interaction participants will report higher levels of social perception of the robot.
   \item \textbf{H2}: Participants’ perception of the robot will be influenced by prior knowledge that they would interact with a robot during the visit.
\end{itemize}
To answer our research questions and to test our hypothesis, we have combined qualitative and quantitative approaches.

\subsection{Method}

Thirty-four Italian participants (Age: $\mu=30.5, \sigma=12.6;$ $17$ females, $17$ males) were invited to join the study through a recruiting mailing list. The research was approved by the Regional Ethical Committee and, as indicated in the informed consent, participants received an appropriate refund for their participation and a free ticket for the museum.
\\ Participants were asked to join the museum for a visit and complete a survey before and after the interaction with the robot (see Section~\ref{selfreported_measurements}). This allowed us to test H1.
\\To test H2 participants were randomly divided into two experimental conditions: 
\begin{itemize}
     
    \item \textit{Announced Condition} (n = 17): Participants were explicitly informed during the sampling phase that they would be interacting with an autonomous museum guide robot.

    \item \textit{Surprise Condition} (n = 17): Participants were only told in the sampling phase that they would take part in an experiment inside a museum, with no mention of the robot.
\end{itemize}

These experimental conditions aimed to simulate real-life scenarios where some visitors expect the robot's presence—due to prior information or promotional materials—while others are surprised to encounter it. To validate this experimental manipulation, all participants were asked about their expectations: those in the "Announced Condition" anticipated the robot, while those in the "Surprise Condition" expected to visit a museum hall.

Upon arrival, participants were introduced to the museum space and the autonomous robot Alter-Ego. After signing informed consent and completing the pre-survey  they began the interaction, which included a structured tour of two mandatory exhibit areas. Participants could then choose to continue the guided tour or end the experiment. The full tour comprised seven areas. 
After the interaction, participants completed the post-survey and were briefly interviewed for feedback.

\section{Measurements}\label{measures} 

\subsection{Self-Reported Measures} \label{selfreported_measurements}
To evaluate participants' impressions of the robot as a social agent and their perception of its technology, they completed pre- and post-experiment surveys. We used established HRI items and administered surveys via the open-source tool SoSci Survey\footnote{\url{https://www.soscisurvey.de/}}. Scales' response format was on a 1-7 point Likert.
\subsubsection{Pre Experiment Survey}
Before the interaction with the robot participants' demographics were collected and they answered the following scales:
\begin{itemize}
    \item \textbf{Previous experiences with robots}: multiple choice questions to understand previous experience with robots and previous experiences with Alter-Ego.
    \item \textbf{Perceived role of the robot} multiple choice questions about the first-impression of the robot ("Alter-Ego for me is like: a toy; a museum assistant; a mechanical component of the museum; a museum installation).
    \item \textbf{Human-like appearance of the robot} \cite{ferrari2016blurring}: a scale on how much the robot was perceived human-like.
    \item \textbf{Agency and Experience} \cite{gray2007dimensions}: a scale to rate if the robot was perceived able to act and feel.
    \item \textbf{Warmth and Competence} \cite{fiske2007universal}: a scale on how much they perceived the robot as competent and friendly.
    \item \textbf{Attitude towards technology }\cite{heerink2009measuring}: a scale on positive attitude towards the robot.
    \item \textbf{IOS} \cite{aron1992inclusion}: single item on the perception of inclusion of the other in the self.
  \end{itemize}
\subsubsection{Post Experiment Survey}
after the interaction participants answered again the Human-like appearance, Agency and Experience, Warmth and Competence and IOS scales.
Moreover, the following measures were added:
\begin{itemize}
    \item \textbf{Evaluation of Robot's Movements}: a scale adapted from \cite{bernotat2023remember} evaluating the movements of the robots during the interaction in terms of fluency and adequacy.
    \item \textbf{Evaluation of Robot's Autonomy}: a scale adapted from \cite{bernotat2023remember} assessing the perceived autonomy and ability of the robot during navigation.
    \item \textbf{Intention of Use} \cite{heerink2009measuring}: a scale to evaluate the participants' willingness to use again the robot in a museum visit.
    \item \textbf{Trust in  Information}: a scale self-generated to evaluate the trustworthiness and accuracy of Alter-Ego when explaining paintings.
    \item \textbf{Perceived Enjoyment} \cite{heerink2009measuring}: a scale to evaluate the participants' enjoyment during the interaction with Alter-Ego.
    \item \textbf{Perceived Sociability}: \cite{heerink2009measuring} a scale assesses the perceived sociability of the robot.
    \item \textbf{Perceived Utility}: a scale adapted from \cite{heerink2009measuring} to evaluate how much participants felt Alter-Ego was useful in the museum visit.
    \item \textbf{Interactiveness}: a self generated scale, to rate how much participant felt the interaction with Alter-Ego interactive. 
\end{itemize}

In the surveys, other items have been collected to evaluate the perception and socio-economic impact of the technology in this context, but they will not be reported in this work. 

\subsection{Behavioral Measurements and Feedbacks}

During the experiments, we have collected the conversation transcriptions (CT) for each participant. The CT presents: 1) the duration of the interaction with the robot 2) the list of exhibition areas visited, and 3) the complete textual conversation.
Due to failures in the robot (i.e., disconnection from the network or battery drained) we were unable to collect the complete CT of three participants.

\section{Data Analysis}

We will now describe the data analysis conducted over the experiments' data. The section is organized as follows: \textbf{1) Qualitative analysis:} authors' annotations about themes emerging both from CT output and from post-experiments feedback.
\textbf{2) Quantitative analysis:} static metrics on the results of the questionnaires and the CT.
\\All statistical analyses have been performed through the open source software Jamovi\footnote{\url{https://www.jamovi.org/}}.

\subsection{Qualitative Analysis}

To address our research question, we conducted an exploratory qualitative analysis using an inductive, bottom-up approach. Rather than applying a predefined coding scheme, we adopted a flexible, data-driven method informed by reflexive thematic analysis \cite{braun2006using}, \cite{braun2019reflecting}. All interaction CTs were thoroughly reviewed by the authors, with emerging themes and notable or unexpected elements annotated in line with a constructivist stance that recognizes the researcher’s active interpretive role. For post-experiment feedback, the most insightful participant comments were noted. This analysis aimed to explore participants’ experiences in greater depth, identifying recurring themes independent of the experimental conditions.
\subsection{Quantitave Analysis}
\subsubsection{Conversation Transcriptions parameters}

By extracting data related to the stops, we then calculated the average frequencies and computed the following parameters:
\begin{itemize}
    
    \item Mean \textbf{duration} of the Interaction with the Robot.
    \item Mean of \textbf{areas visited} in the exhibition after the two mandatory ones.
    \item Mean of \textbf{questions} to the robot during the visit.
    \item Mean of \textbf{answers} by the robot during the visit: this parameter included only the correct answers provided by the robot, meaning instances in which the robot was able to respond to participants in a satisfactory manner (e.g., Participant (P)19: \textit{''Which type of ship is represented in this painting?''} Robot: \textit{''It is a military cruiser''}).
    \item Mean of \textbf{out-of-scope questions} during the visit: this parameter refers to the instances in which the robot explicitly stated it did not have the requested information (e.g., P23: "\textit{What is the most beautiful ocean liner ever built?}" Robot: "\textit{I’m not aware of this information. You should ask the museum staff}").
    \item Mean of \textbf{comprehension failures} by the robot (e.g."\textit{Could you repeat your question? I didn’t understand}").
    
\end{itemize}
\subsubsection{Self-Reported Measures}
After reverse-coding relevant items, Cronbach’s $\alpha$ was calculated to assess scale's internal consistency. We then computed mean values for \textit{Experience with robots}, \textit{ with Alter-Ego}, and \textit{Perceived role of Alter-Ego} to characterize the sample. Data normality was tested using the Shapiro-Wilk test, guiding the choice of statistical tests: non-parametric (Wilcoxon, Mann-Whitney U) when normality was violated, and parametric (paired and independent t-tests) when it was met. To test H1, we compared pre- and post-interaction self-reports (Agency, Experience, Warmth, Competence, IOS). To test H2, we compared these measures across experimental conditions.



\begin{figure}[htbp] 
    \centering  
    \includegraphics[width=0.5\textwidth, scale=0.10]{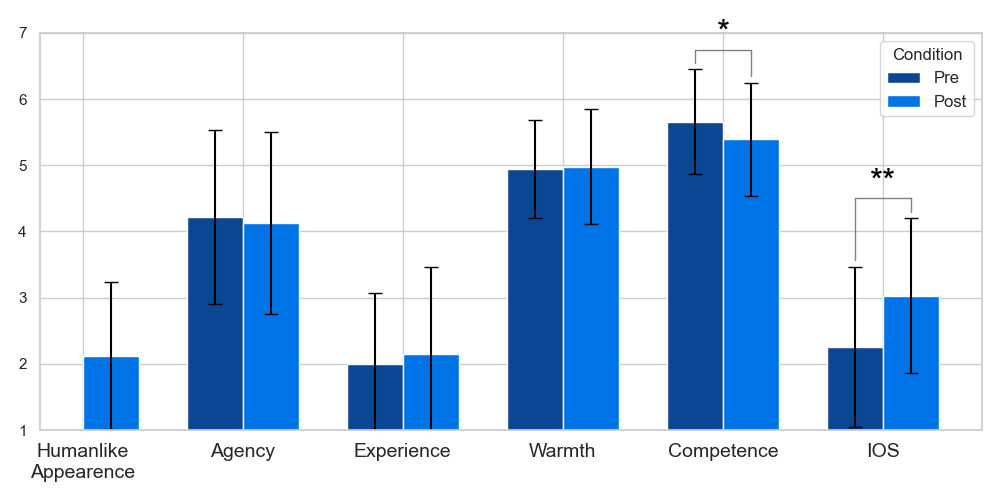}  
    \caption{Mean Likert scores (1–7 scale) from pre- and post-experiment surveys. The pre-experiment score for Human-like appearance is excluded due to a Cronbach's alpha below acceptance rate.Asterisks indicate significance levels: p < .05 (*), p < .01 (**), for differences between pre- and post-experiment scores.} 
    
    \label{fig:prepost}  
\end{figure}

\section{Results} \label{results}
\subsection{Qualitative Analysis}

We conducted an exploratory qualitative analysis to examine participants’ interactions and feedback, identifying five key themes that reflect both user strategies and system limitations. This analysis was inspired by a reflexive thematic approach \cite{braun2021reflexive, nowell2017thematic} and was grounded in a constructivist perspective.

\subsubsection{\textbf{Modes of Verbal Interaction}}
Participants displayed a range of verbal behaviors, from testing the robot’s sociality through personal or abstract questions (e.g., “\textit{What do you think?}”, P09; “\textit{Who’s your favorite painting?}”, P13), to using humor or imitating its speech style (P23). Some aimed to naturalize the exchange (P02, P13), while others requested repetition or clarification (e.g., “\textit{Can you remind me of your name?}”, P07). A noteworthy instance involved a participant (P13) asking to be addressed using gendered language, despite the robot employing neutral forms.

\subsubsection{\textbf{Attitudes, Emotions, and Engagement}}
Initial engagement was generally high, with participants expressing curiosity and amusement (P06, P13). However, this enthusiasm sometimes gave way to frustration, especially when expectations were unmet or technical limitations emerged (P11, P29). Perceived failures in comprehension led some participants to search for implicit cues in the robot’s behavior (e.g., P07 interpreting arm gestures as feedback signals). Despite occasional issues, most participants reported overall satisfaction and expressed willingness to re-engage in the future (P15, P28).

\subsubsection{\textbf{Expectations}}
Participants often projected high expectations onto the robot, comparing it to advanced systems like ChatGPT (P07) or other robots such as iCub \cite{metta2010icub} (P29). Disappointment was voiced when the system failed to meet perceived standards of fluency or spontaneity (P11), or lacked domain-specific knowledge (P09). Some assumed the robot could infer internal states, such as confusion, suggesting anthropomorphic attributions (P14).

\begin{figure}[htbp] 
    \centering  
    \includegraphics[width=0.5\textwidth, scale=0.10]{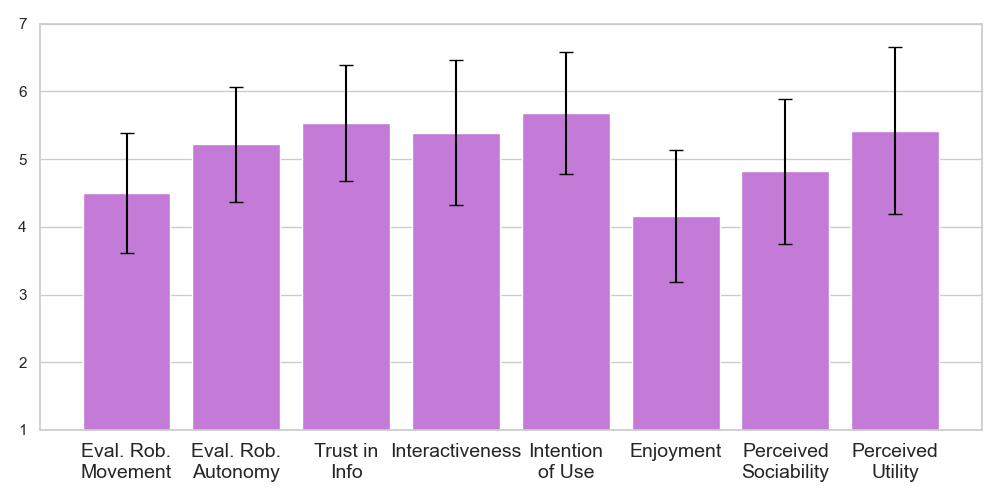}
    \caption{Mean Likert scores (1–7 scale) from the post experiment survey evaluating the robot.}  
    \label{fig:robot-perc}  
\end{figure}

\subsubsection{\textbf{Linguistic and Contextual Challenges}}
Numerous interaction breakdowns stemmed from the robot’s limited comprehension and contextual disambiguation. Participants struggled when referring to specific stops, often repeating or rephrasing inputs (P03, P15, P21), while one eventually abandoned the interaction (P08). Latency in responses was frequently mentioned as a source of frustration (P07, P29). These issues highlight ongoing challenges in aligning speech recognition with contextual awareness in dynamic, real-world settings \cite{goodrich2008human}.

\subsubsection{\textbf{Control and Decision-Making Autonomy}}
Participants exhibited varying degrees of agency when navigating the tour. While some delegated decisions to the robot (P13, P27), most maintained control over the experience. In two cases (P05, P16), the tour was unintentionally ended due to ambiguous phrasing, indicating misinterpretation by the system. Notably, P26 requested to pause the robot-led tour to explore independently, later re-engaging—a behavior not anticipated by the system and unrecognized as intentional, pointing to the need for improved handling of implicit user intent.

\subsection{Quantitative Analysis: Parameters of the visit}


In table \ref{tab:ST parameters} we report the values of the parameters extracted from the visit.
\begin{table}[h]
\centering
\caption{}
\label{tab:ST parameters}
\begin{tabular}{|l|l|l|}
\hline
\textbf{Parameter}              & \textbf{$\mu$} & \textbf{$\sigma$} \\ \hline
\begin{tabular}[c]{@{}l@{}}Duration of Interaction  (minutes)\end{tabular} & 19.20 & 18.5 \\ \hline
Areas Visited           & 5.69 & 1.79 \\ \hline
Questions              & 7.29 & 6.21 \\ \hline
Answers                & 3.71 & 3.69 \\ \hline
Out-of-scope questions             & 2.25 & 1.8  \\ \hline
Comprehension failures & 1.26 & 1.35 \\ \hline
\end{tabular}
\end{table}


Among the observed interactions, participants asked a mean of 7.29 questions (SD = $6.21$), of which 3.71 (SD = $3.69$) fell within the scope of the robot’s knowledge base, allowing it to provide appropriate responses. Questions outside this scope typically concerned artworks or details not covered by the museum-provided data (e.g. "How fast was this ship?").
To assess the robot’s performance across different exhibit areas, we computed error rates as the proportion of misunderstandings per visitor. Specifically, the number of comprehension failures in each area was divided by the total number of visitors to that area.
The highest error rates were observed in the Military Ships and Emigration areas (both at 33.33\%), which were also the noisiest, due to ambient sounds and audio installations present in the exhibits. In contrast, the quieter Port of Genoa had the lowest error rate (5.88\%).

\subsection{Quantitivative Analysis: descriptives, perception of the robot and the visit}

All the multiple-item scales demonstrated good internal consistency in both pre and post administrations (all $\alpha$ $>$ $0.70$) \cite{nunnally1978psychometric}, except for the Human-like Appearance scale in the Pre Survey ($\alpha$=0.33), which was then discarded by the analysis. 
The mean for previous experience with robots was $2.41$ (SD = $0.67$) on a 5-point scale, while the attitudes towards technology scale had a mean of $5.23$ (SD = $0.89$).
Most participants ($67.6$\%) reported not having seen the robot before. Some participants had seen the robot on social media, TV, or in the lab, but none had previously interacted with Alter-Ego in an experiment. 


Answering our \textbf{RQ}, Figure 4 shows the mean values of the scales to the general perception of the robot, all above the midpoint, indicating a generally positive perception.
Regarding \textbf{H1}, in figure \ref{fig:prepost} we report the results of the pre- and post-comparisons on the dependent variables investigating the social perception of the robot.
The participants' perception of robot competence differed significantly between the two measures ($\mu_{pre}$=$5.66$, $\sigma_{pre}$=$5.39$, $\mu_{post}$=$5.39$, $\sigma_{post}$=$0.86$ $p = .019$), suggesting a slight worsening in perceived competence. A significant difference also emerged in the IOS (Inclusion of Other in the Self) scores ($\mu_{pre}$=$2.26$, $\sigma_{pre}$=$1.21$, $\mu_{post}$=$3.03$, $\sigma_{post}$=$1.17$, $p = .003$), indicating a greater closeness with the robot after the interaction. No significant differences were observed for Experience, Agency, or Warmth. Regarding \textbf{H2} we check the effect of the condition on the variables. We found no significant effect of the condition over any of the self-reported variables expect for the variable "means of areas visited". The test showed a marginally significant difference ($p = 0.053$), indicating that the scores in condition Announced tend to be lower ($\mu_{A}= 5.24$, $\sigma_{A}= 1.56$) than those in condition Surprise ($\mu_{S}= 6.20$, $\sigma_{S}= 1.93$). 


\section{Conclusion}

This study offers empirical evidence on how users perceive and relate to a socially interactive robot across cognitive, affective, and relational dimensions. Through a mixed-method design, we observed both measurable and interpretative shifts in perceived agency, competence, and closeness after interaction. Findings suggest that embodiment, interaction dynamics, and contextual framing critically shape social responses to artificial agents.

Our results highlight the importance of a multi-method approach in HRI, revealing not only statistical trends but also participants’ situated experiences. Framing the robot as a social alter ego—rather than a neutral tool—emerged as a key factor in user engagement.

Future research should examine long-term effects, user diversity, robot behavior variations, and how transparency, autonomy, and explainability influence trust. 

\addtolength{\textheight}{-12cm}   





\section*{ACKNOWLEDGMENT}
This work was supported by:
"RAISE – Robotics and AI for Socioeconomic Empowerment" and by the European Union - NextGenerationEU.
Thanks to Giovanni Rosato, Eleonora Sguerri, Mattia Poggiani, Cristiano Petrocelli and Manuel Barbarossa for the technical support.


\bibliographystyle{IEEEtran} 
\bibliography{IEEEabrv,bibliography}

\end{document}